\documentclass{article} 
\usepackage{iclr2027_conference,times}
\usepackage{xspace}
\DeclareRobustCommand{\method}{TGR-EvoDAG\xspace}

\usepackage{amsmath,amsfonts,bm}

\def\eqref#1{equation~\ref{#1}}

\def\1{\bm{1}}

\DeclareMathAlphabet{\mathsfit}{\encodingdefault}{\sfdefault}{m}{sl}
\SetMathAlphabet{\mathsfit}{bold}{\encodingdefault}{\sfdefault}{bx}{n}

\usepackage{url}
\usepackage[T1]{fontenc}        
\usepackage[utf8]{inputenc}
\usepackage{microtype}
\usepackage{hyperref}
\hypersetup{
    colorlinks=true,      
    linkcolor=blue,       
    citecolor=blue,       
    urlcolor=blue         
}
\usepackage{url}
\usepackage{xurl}
\usepackage{graphicx}
\usepackage{multicol}
\usepackage{multirow}
\usepackage{array}
\usepackage{amsmath,amssymb,bm}
\usepackage{booktabs}
\usepackage{enumitem}
\usepackage{wrapfig}
\usepackage{algorithm}
\usepackage{algorithmic}        
\usepackage{colortbl}
\usepackage{caption}
\usepackage{subcaption}
\usepackage{xcolor}
\usepackage{setspace}
\usepackage{tabularx}
\usepackage{blindtext}
\usepackage{pgfplots}
\pgfplotsset{compat=1.18} 
\usepackage{tikz}
\usetikzlibrary{er,positioning,bayesnet}
\usepackage{makecell}
\usepackage{siunitx}
\usepackage{nicefrac}
\usepackage{tocloft}
\usepackage{listings}
\usepackage{tcolorbox}
\tcbuselibrary{skins,breakable}
\usepackage{xltabular}
\usepackage{adjustbox}
\usepackage{lipsum}
\usepackage{graphicx}
\usepackage{caption}
\usepackage{booktabs}

\usepackage{listings}

\newif\ifshowtodo
\showtodofalse

\ifshowtodo
  \newcommand{\TODO}[1]{%
    \textcolor{red}{\textbf{TODO:} #1}%
  }
  \newcommand{\CTODO}[1]{%
    \TODO{#1}%
  }
\else
  \newcommand{\TODO}[1]{}
  \newcommand{\CTODO}[1]{}
\fi

\title{TGR-EvoDAG: A DAG-based Multi-Agent Code Evolution Framework for Generative Recommendation on Tencent}

\author{
  \parbox{\linewidth}{\centering\normalfont
    {\bfseries Junlin He}\textsuperscript{1,2,*}, \quad
    {\bfseries Junzhang Zhu}\textsuperscript{1,*}, \quad
    {\bfseries Yudong Li}\textsuperscript{1,*}, \quad
    {\bfseries Yihong Tang}\textsuperscript{3}, \\
    {\bfseries Xuefeng Jiang}\textsuperscript{4}, \quad
    {\bfseries Wei Ma}\textsuperscript{2}, \quad
    {\bfseries Chengxiang Zhuo}\textsuperscript{1}, \quad
    {\bfseries Zang Li}\textsuperscript{1} \\[2ex]
    \small
    \textsuperscript{1}Platform and Content Group, Tencent, Shenzhen, China \\
    \textsuperscript{2}The Hong Kong Polytechnic University, Hong Kong, China \\
    \textsuperscript{3}McGill University, Montreal, Canada \qquad
    \textsuperscript{4}Independent Researcher, China
  }
}

\iclrfinalcopy
\begin{document}

\maketitle
\lhead{Preprint.}

\begingroup
\renewcommand{\thefootnote}{\fnsymbol{footnote}}
\footnotetext[1]{Equal contribution.}
\endgroup

\begin{abstract}
Large Language Model (LLM) agents have enabled automated code evolution, offering a promising paradigm for algorithm discovery. However, current recursive self-improvement (RSI) research for sequential recommendation faces a critical dilemma: general-purpose code evolution frameworks possess flexible search mechanisms but lack domain-specific feedback, whereas recommender-specific evolution frameworks utilize rich feedback but rely on linear evolution. More critically, naively integrating them fails when applied to Generative Recommendation (GR). Unlike traditional end-to-end models, GR relies on a module-cascaded training pipeline (i.e., tokenization, pre-training, and alignment). In this paradigm, existing naive evolution methods inevitably suffer from \emph{i) cost blindness:} squandering budgets by ignoring stage-asymmetric retraining overheads; and \emph{ii) short-sighted evaluation:} discarding promising upstream innovations that temporarily regress metrics before downstream adaptation. To bridge these gaps, we propose \method, a multi-agent code evolution framework explicitly tailored for GR. \method organizes code evolution as a Directed Acyclic Graph (DAG) that explicitly maintains multiple evolution branches and merges complementary innovations, and harmonizes an advanced search mechanism with rich GR-specific diagnostic feedback. We further introduce \emph{stage-aware cost planning} to optimize real-world GPU-hour budgets, and \emph{active downstream co-adaptation} to navigate temporary regressions and unlock deferred cross-stage gains. On public Amazon datasets, \method marks the first autonomous evolution of a complete GR pipeline, improving offline metrics by up to 37.4\% and boosting Agent4Rec simulated user satisfaction by 19.7\% over the strongest GR baseline. Furthermore, in industrial settings, the evolved GR achieves a 12.16\% relative improvement in offline Hit@1 and, when deployed as an additional retrieval channel on Tencent's large-scale commercial media platform, a statistically significant 0.40\% increase in online click-through rate (CTR).
\end{abstract}

\section{Introduction}

With the emergence of Large Language Model (LLM) agents, LLM-driven code evolution has emerged as a promising paradigm, paving the way for Recursive Self-Improvement \citep{romera2024funsearch,jiang2025aide,chen2026recursiveselfimprovementaibounded}. This approach gives agents research-like capabilities: they propose hypotheses, 
modify code, run experiments, and collect feedback, enabling algorithms to 
self-evolve continuously \citep{lu2024aiscientist,chan2025mlebench,huang2024mlagentbench,liu2026diveclaudecodedesign,chen2021evaluatinglargelanguagemodels}. To accelerate experimental iterations and reduce labor costs for recommendation engineers, an increasing number of studies have introduced this paradigm into the field of sequential recommendation \citep{wang2026self,kim2026self,mu2026evorec}.

While general-purpose code evolution has achieved sophisticated search mechanisms, such as multi-branch exploration and cross-branch merging, through population-based approaches \citep{romera2024funsearch,lange2025shinkaevolve,jiang2025aide}
and tree/graph-based architectures \citep{jin2026toward}, they generally lack the domain-specific feedback signals essential for recommender systems. Conversely, existing recommender evolution frameworks invest heavily in feedback quality by utilizing simulated users and diagnostic probes, yet they predominantly rely on primitive search mechanisms. For instance, SERS \citep{wang2026self} uses a simple fan-out/fan-in strategy without branch maintenance; EvoRec performs greedy, round-by-round linear advancement \citep{mu2026evorec}; and Self-EvolveRec samples a parent node from multiple elites \citep{kim2026self}.

These simple search mechanisms restrict existing recommender evolution frameworks to traditional end-to-end models, where modifications incur a uniform full-training cost and produce immediate metric changes \citep{kang2018sasrec,hidasi2016gru4rec,sun2019bert4rec}. They \textbf{fail on Generative Recommendation (GR)}, an increasingly popular paradigm 
built on a cascaded pipeline (tokenizer, pre-training, and alignment)\citep{geng2022p5,rajput2023tiger,wang2024letter,deng2025onerec,pang2025higr,tgrteam2026tgradvancingindustrialrecommendation}, because 
they cannot effectively maintain or merge cross-stage innovations.
While explicitly combining advanced general-purpose search mechanisms with domain-specific feedback seems like a natural solution, this naive integration still falls short. General-purpose mechanisms fundamentally rely on two implicit assumptions that a cascaded GR pipeline structurally violates:
\textbf{1) Cost Blindness vs. Stage-Dependent Costs.} General-purpose search policies evaluate all candidate nodes uniformly, ignoring \textit{how much retraining} an edit requires. In GR, however, retraining costs are highly asymmetric: modifying the upstream Tokenizer forces a full-pipeline retraining, whereas editing the downstream Alignment stage only re-runs the final segment. Being ``cost-blind,'' existing frameworks cannot triage edits based on actual overhead, systematically squandering budgets on expensive upstream explorations.
\textbf{2) Short-Sighted Evaluation vs. Deferred Gains.} General-purpose frameworks judge modifications by their immediate scores, pruning candidates that initially regress performance. However, because GR stages are tightly coupled, modifying an upstream stage almost always breaks compatibility with the unadapted downstream stage, causing a temporary performance drop. Lacking structural awareness, existing frameworks abandon these upstream innovations, missing the opportunity to actively schedule the necessary downstream adaptations to unlock ``deferred'' gains\citep{liu2025etegrec}.

\begin{figure*}[t]
    \centering
    \includegraphics[width=0.95\textwidth]{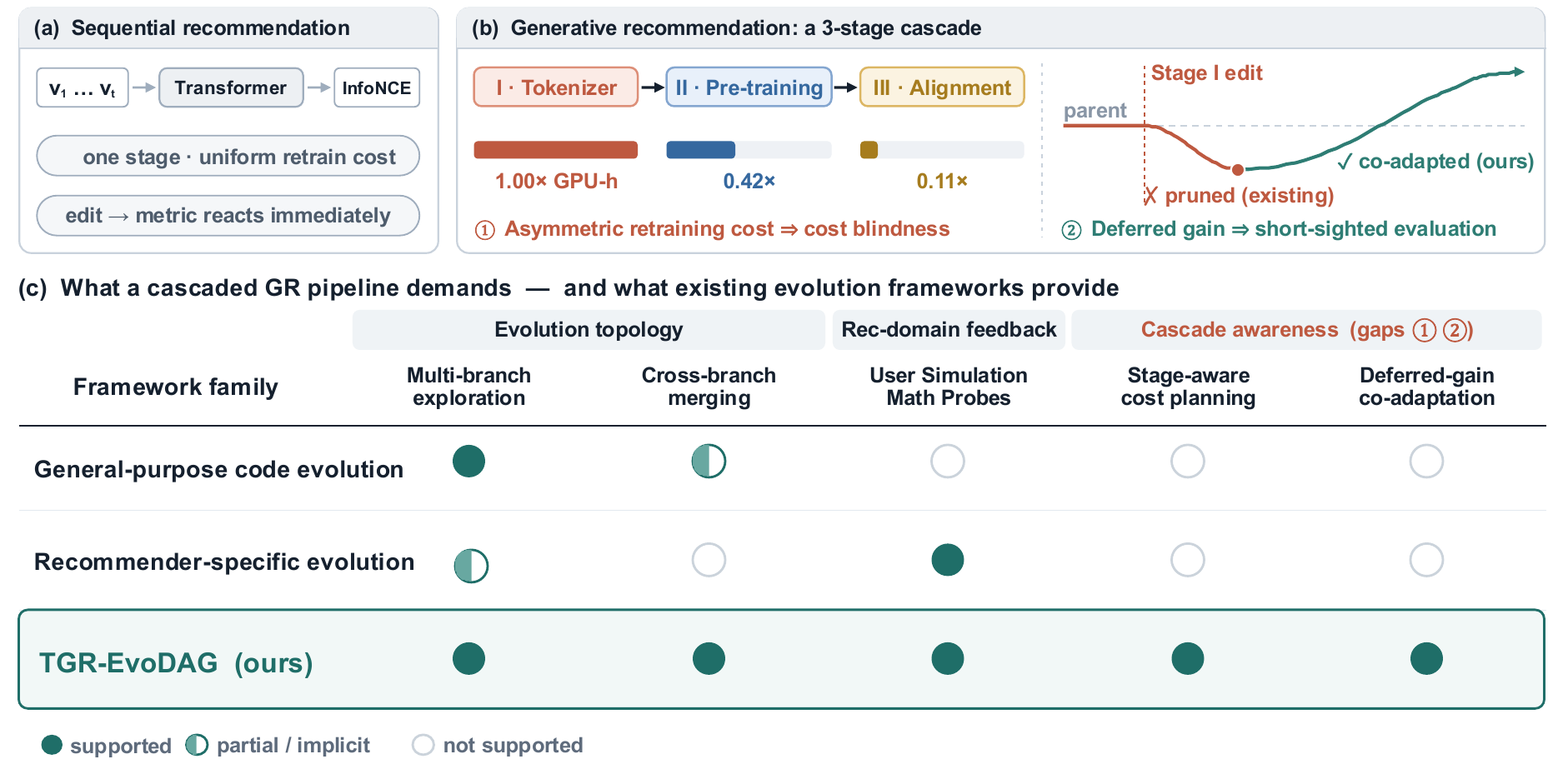}
    \caption{Comparison of recommendation paradigms and evolution framework capabilities. (a) Traditional sequential recommendation relies on an end-to-end architecture characterized by uniform retraining costs and immediate metric feedback. (b) GR utilizes a cascaded pipeline, which introduces stage-asymmetric retraining costs and deferred cross-stage gains, causing existing frameworks to suffer from \emph{cost blindness} and \emph{short-sighted evaluation}. (c) Capability comparison demonstrating that \method uniquely fulfills the specific demands required to evolve cascaded GR pipelines.}
    \label{fig:motivation}
    \vspace{-2mm}
\end{figure*}

To this end, we present \method, a multi-agent code evolution framework explicitly built for cascaded GR pipelines. \method seamlessly marries advanced graph search mechanisms with rich domain-specific feedback. It represents the search space as a Code Directed Acyclic Graph (Code DAG), driven by a five-stage collaborative Agent workflow (comprising an Operator Selector, Architect-Critic, Coder-Reviewer, Distiller, and Pruner) that digests fine-grained recommendation signals, such as mathematical probes and simulated user  \citep{zhang2024agent4rec}. Building upon this foundation, \method tightly integrates the multi-agent workflow with two novel mechanisms to overcome the inherent blind spots of cascaded pipelines.
First, to address cost blindness, we introduce \textbf{Stage-Aware Cost Planning via the Operator Selector.} Each DAG node records the exact pipeline stages its edit touches and the corresponding retraining overhead. Empowered by this stage-aware schema, our Operator Selector dynamically balances the expected performance gain against the actual GPU-hour cost, preventing the framework from blindly exhausting budgets on expensive upstream explorations. Second, to overcome short-sighted evaluation, we design \textbf{Active Downstream Co-Adaptation.} Instead of rigidly discarding candidates that cause immediate performance drops, \method explicitly retains these seemingly ``regressed'' upstream branches. When the framework identifies an upstream edit (e.g., Tokenizer) lacking a corresponding downstream adaptation (e.g., Pre-training), the agents deliberately schedule a child node to adapt the downstream stage. This recovers gains from upstream edits that greedy frameworks would discard.

To evaluate our framework fairly, we construct a standardized GR evolution platform equipped with stage-aware feedback and precise GPU-hour accounting, ensuring all compared frameworks evolve from an identical seed codebase. In summary, our core contributions are:

\begin{itemize}[leftmargin=*] 
\item We identify and formulate the structural mismatch between current code evolution frameworks and cascaded GR pipelines. Specifically, we expose how cost blindness and short-sighted evaluation cause existing methods to systematically misallocate computational budgets and discard promising cross-stage innovations.
\item We propose \method, which harmonizes Code DAG search mechanisms with domain-specific feedback via a comprehensive five-stage multi-agent workflow. By further introducing stage-aware cost planning and active co-adaptation, \method enables efficient GR evolution.
\item We develop a standardized GR evolution platform. Extensive experiments demonstrate that \method achieves the fully autonomous evolution of a complete GR training pipeline, surpassing both general and recommender-specific evolutionary baselines. It improves offline metrics by up to 37.4\% and boosts user satisfaction in the Agent4Rec simulated environment by 19.7\% over the 
strongest GR baseline. Moreover, applying the evolved architecture to \textbf{Tencent's real-world industrial scenarios} yields a 12.16\% relative improvement in offline Hit@1 and a statistically significant 0.40\% increase in online Click-Through Rate in a 14-day A/B test, demonstrating its practical value.
\end{itemize}

\method is part of the TGR series of generative recommendation research at Tencent~\citep{tgrteam2026tgradvancingindustrialrecommendation}. Beyond this paper, \method has been integrated into Tencent's internal Uppimo platform as its implementation backbone, which productizes the recursive self-improvement (RSI) workflow described here for production recommendation teams; the platform will be made available for public trial in the future.

\section{Related Work} \label{sec:related-work}

\smallskip \noindent\textbf{LLM-driven Code Evolution.}
General-purpose code evolution relies on three search paradigms. 
\textbf{Linear inheritance} advances a single thread (e.g., AIDE, MLE-STAR~\citep{nam2025mlestar,yang2025rdagent,guo2024dsagent}), often stalling at local optima. 
\textbf{Multi-elite sampling} draws parents from elite archives to preserve diversity~\citep{romera2024funsearch,novikov2025alphaevolve,lange2025shinkaevolve,liu2025deepevolve,liu2024eoh,ye2024reevo}. 
\textbf{Explicit tree search} maintains multiple branches for persistent exploration (e.g., SELA, Arbor~\citep{chi2024sela,jin2026toward}); notably, CORAL~\citep{qu2026coral} even removes fixed strategies and delegates all search routing to the agent itself. 
Despite their sophisticated search, these frameworks fail on Generative Recommendation (GR). First, they lack the domain-specific feedback (e.g., mathematical probes, simulated users) essential for GR. Second, facing GR's cascaded pipeline, they lack stage-aware cost planning to handle asymmetric retraining overheads, and cross-stage mechanisms to merge and co-adapt innovations from upstream tokenization down to alignment.

\smallskip \noindent\textbf{Generative Recommendation (GR).}
GR replaces classical ID-based models~\citep{hidasi2016gru4rec,kang2018sasrec,sun2019bert4rec} with direct semantic-ID generation~\citep{tay2022dsi,rajput2023tiger}. Training GR forms a \emph{cascaded three-stage pipeline}: a Tokenizer quantizes items into hierarchical IDs~\citep{lee2022rqvae}; Pre-training fits a sequence-to-sequence generator; and Post-training aligns it with preference signals~\citep{chen2024sdpo,deng2025onerec,zhou2025onerecv2,Yu_2026}. Current manual research optimizes each stage locally, spanning better tokenizers~\citep{zheng2024lcrec,wang2024letter,tan2024idgenrec,hou2025actionpiece}, quantization~\citep{fifty2025rotation}, and architectures~\citep{zhai2024hstu}. Because stage-wise improvements are \emph{complementary}, the global optimum hides in their combinations, a space no single-threaded search can cover. Furthermore, these stages are tightly coupled~\citep{liu2025etegrec}; upstream innovations often trigger temporary metric drops, requiring downstream re-adaptation to unlock their \emph{delayed} value. No prior work autonomously evolves this complete pipeline.

\smallskip \noindent\textbf{Self-Evolving Recommender Systems.}
A nascent line applies code evolution to recommenders.
SERS
\citep{wang2026self}, deployed in YouTube's production system, evolves delta edits
against a fixed baseline via a simple fan-out/fan-in strategy. EvoRec
\citep{mu2026evorec} advances the model architecture greedily, round by round,
keeping only the single best variant at each step. Self-EvolveRec
\citep{kim2026self} employs an LLM-based user simulator
\citep{park2023generative,zhang2024agent4rec,wang2025recagent}, and samples a
single parent from multiple elites.
These frameworks all introduce recommender-specific feedback signals, yet
adopt a simple search mechanism that maintains and merges no parallel branches.
They fail when evolving a cascaded GR pipeline: its complementary cross-stage gains
must be \emph{merged} to reach the global optimum, and its upstream edits pay off only
after downstream re-adaptation.
Concurrently, Auto-RecSys~\citep{li2026autorecsys} applies autonomous research agents to industry-scale recommender models. It focuses on engineering orchestration for long-running experiments (persistent state, failure recovery, and human-in-the-loop checkpoints), which is complementary to our work; its search advances each idea linearly, without explicit GPU-hour budgeting, stage-aware cost planning, or cross-branch merging.

\section{Preliminaries} \label{sec:problem}

Let $\mathcal{U}$ and $\mathcal{I}$ denote the sets of users and items. In sequential recommendation, given a user's chronological interaction history $S_u$, the objective is to predict the next item $i_{t+1}$. Generative Recommendation (GR) reformulates this as a conditional sequence generation problem. It maps items to discrete semantic identifiers (SIDs) and autoregressively generates the SID tokens of $i_{t+1}$ conditioned on $S_u$. Training a GR model therefore involves three tightly coupled stages:
(1) \textbf{Stage I: Tokenization.} A tokenizer (e.g., RQ-VAE) discretizes item content into hierarchical semantic IDs.
(2) \textbf{Stage II: Generative Pre-training.} A sequence-to-sequence Transformer is trained to predict the next item's tokens via hierarchical cross-entropy.
(3) \textbf{Stage III: Post-training Alignment.} The model is further aligned with desired recommendation objectives (e.g., exact match or prefix similarity) using preference optimization or reinforcement learning.
We detail the concrete implementation details and mathematical formulations of these three stages (e.g., RQ-Kmeans quantization, exact baseline objectives, and post-training variants) to Appendix~\ref{appendix:codebase}.

\section{\method} \label{sec:methodology}
\begin{figure}[h]
\vspace{-2mm}
\centering
\includegraphics[width=0.8\columnwidth]{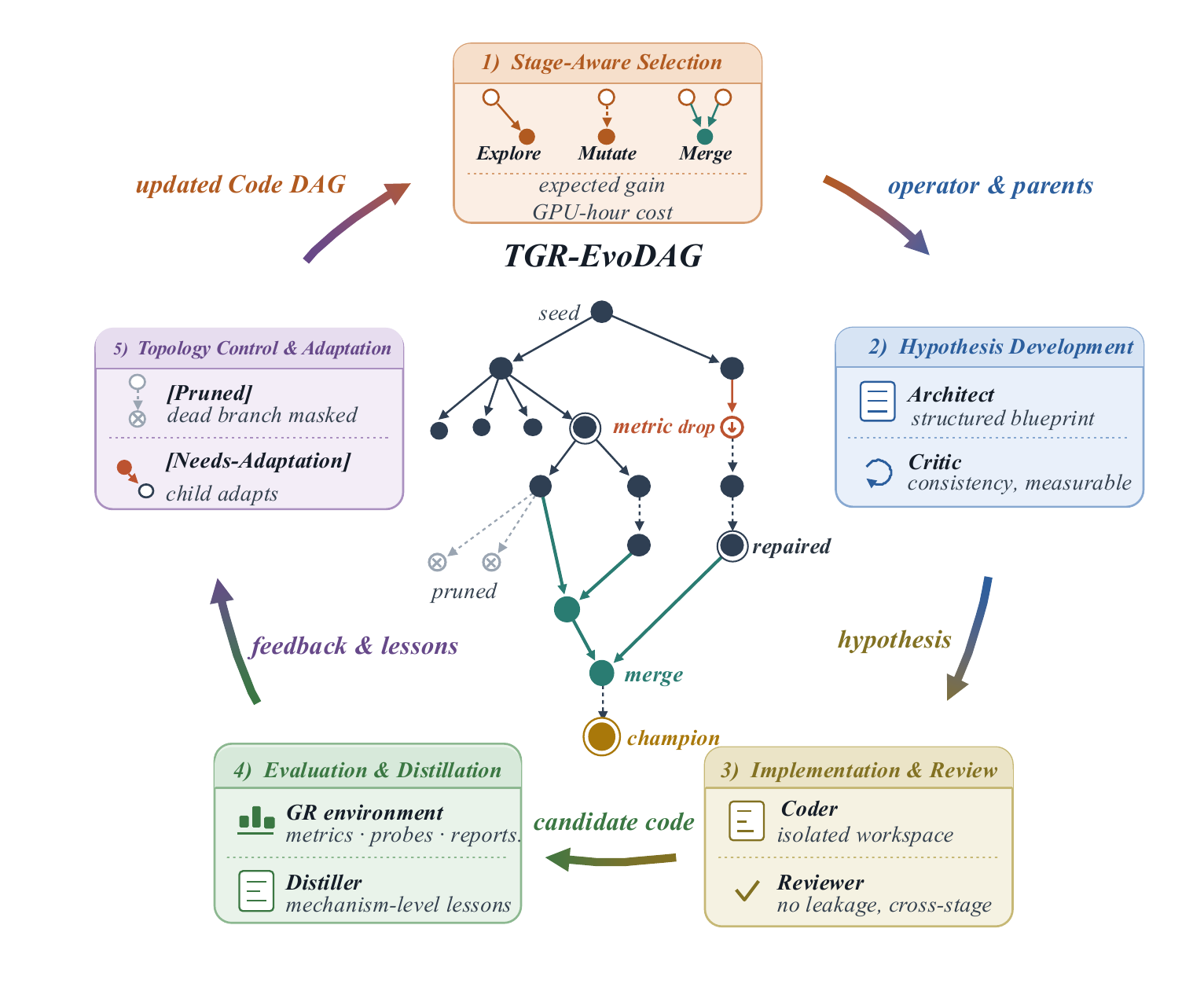}
\vspace{-2mm}
\caption{Workflow of \method.}
\label{fig:workflow}
\vspace{-2mm}
\end{figure}

We present \textbf{\method}, a multi-agent code evolution framework designed for the cascaded training pipeline of generative recommendation. As shown in Figure~\ref{fig:workflow}, \method explicitly represents the evolution history as a Code DAG. This representation enables the framework to preserve multiple promising branches, revisit earlier pipeline stages, and semantically merge complementary improvements discovered along different branches.

At evolution round $t$, the global search state is encapsulated as $\mathcal{S}^{(t)} = (\mathcal{G}^{(t)}, b^{(t)})$, where $\mathcal{G}^{(t)}=(\mathcal{V}^{(t)},\mathcal{E}^{(t)})$ is the current Code DAG, and $b^{(t)}$ denotes the remaining evolution budget. Each node $v_i\in\mathcal{V}^{(t)}$ acts as a self-contained experimental record tracking three categories of structured attributes: (1) \textbf{Provenance}, including the parent set $\mathcal{P}_i$, applied search operator $o_i \in \{ \textsc{Explore}, \textsc{Mutate}, \textsc{Merge} \}$, underlying hypothesis $\mathcal{H}_i$, and branch status $\mathcal{Q}_i$; (2) \textbf{Artifacts}, covering the incremental code modification $\Delta_i$, executable codebase snapshot $\mathcal{B}_i$, and associated artifacts $\mathcal{A}_i$ (e.g., vocabularies, checkpoints); and (3) \textbf{Empirical Execution}, storing the validation feedback $\mathcal{F}_i$, distilled lessons $\mathcal{D}_i$, and consumed experimental cost $\mathcal{C}_i$. The edge set is naturally induced by developmental dependencies, $\mathcal{E}^{(t)} = \{ (u, v_i) \mid v_i \in \mathcal{V}^{(t)}, u \in \mathcal{P}_i \}$. Because each new node is derived exclusively from previously evaluated nodes and never modifies its ancestry, $\mathcal{G}^{(t)}$ is strictly guaranteed to be a Directed Acyclic Graph.

\subsection{Search Operators} \label{sec:dag-ops}

\method transitions the search state by applying one of three search operators to generate a new child node $v_c$. These operators determine the parent set $\mathcal{P}_c$ and dictate the semantic instructions delivered to the multi-agent system.

\paragraph{Exploration.} 
Exploration branches out from a single parent node ($\mathcal{P}_c = \{v_p\}$) to investigate a fundamentally different direction. Agents formulate a novel hypothesis $\mathcal{H}_c$ distinct from its ancestry, such as introducing a new training objective, changing the user-conditioning architecture, or applying a different post-training algorithm.

\paragraph{Mutation.} 
Mutation performs exploitation and local refinement on a single parent node ($\mathcal{P}_c = \{v_p\}$). Agents keep the core hypothesis of $v_p$ intact while performing localized optimizations, such as adjusting loss weights, tuning hyperparameters, or modifying sampling strategies. 

\paragraph{Merge.} 
The \textsc{Merge} operator synthesizes complementary improvements discovered across multiple independent branches ($\mathcal{P}_c = \{v_{p_1}, \dots, v_{p_r}\}$, $r \geq 2$). Unlike single-parent operators, \textsc{Merge} aggregates a multi-dimensional context. The framework packages the hypotheses, codebases, and validation feedback of all parent nodes and forwards them to the pipeline. Agents are tasked to identify complementary mechanisms, resolve semantic code conflicts or interface dependencies, and integrate them into a unified functional codebase $\mathcal{B}_c$ under a synergized hypothesis $\mathcal{H}_c$. To keep the context tractable, we do not inject full codebase snapshots; instead, each parent is serialized as a unified diff against the lowest common ancestor in the DAG. Each diff is paired with its structured hypothesis $\mathcal{H}$ and distilled lessons $\mathcal{D}$. This representation stays within the context window while preserving the information needed to resolve cross-branch conflicts.

\subsection{Multi-Agent Pipeline} \label{sec:agents} 

Each node expansion is executed by a five-stage pipeline instantiated by seven specialized LLM agents: an Operator Selector, an Architect-Critic pair, a Coder-Reviewer pair, a Distiller, and a Pruner. To ensure rigorous evolution, agents communicate through structured intermediate records (e.g., JSON schemas) rather than unconstrained natural language. The complete workflow is summarized in Algorithm~\ref{alg:evodagrec} (Appendix~\ref{appendix:workflow}).

\paragraph{Stage-Aware Cost Planning via Operator Selector.}
Instead of simply counting search iterations, \method manages a realistic GPU-hour budget $b^{(t)}$ via empirical cost profiling. Every evaluated node explicitly records the actual time spent on each pipeline stage (e.g., Tokenizer, Pre-training, Alignment). Provided with these historical records, the Operator Selector specifies the \texttt{target\_stage} and \texttt{training\_scope} of its intended edit using a structured JSON schema. This allows the system to estimate the expected GPU-hour cost $\mathcal{C}_c$ before execution. By weighing this expected cost against anticipated performance gains, the Operator Selector makes strategic budgeting decisions. For instance, it deliberately postpones expensive upstream edits (which force full-pipeline retraining) until cheaper downstream architectures fully mature, preventing premature budget exhaustion.

\paragraph{Hypothesis development.}
Conditioned on the selected operator $o_c$ and parents $\mathcal{P}_c$, the Architect formulates the hypothesis $\mathcal{H}_c$ as a structured blueprint detailing the diagnosed limitation, proposed mechanism, affected stages, expected diagnostic changes, and potential risks. The Critic then evaluates this blueprint against historical lessons and graph ancestry to verify logical consistency and ensure measurable effects. They engage in an iterative refinement loop until the hypothesis passes all structural and semantic checks.

\paragraph{Implementation and review.}
Guided by $\mathcal{H}_c$, the Coder implements the codebase snapshot $\mathcal{B}_c$ in an isolated workspace initialized from the parents. The Coder and Reviewer execute a multi-round repair loop informed by static analysis and peer review. The Reviewer verifies that the implementation faithfully realizes the blueprint, preserves the evaluation protocol without data leakage, and maintains cross-stage compatibility. For \textsc{Merge} nodes, it additionally ensures mechanisms inherited from all branches remain functionally active. This loop terminates when the candidate passes all tests or exhausts the repair budget.

\paragraph{Evaluation and lesson distillation.}
The standardized environment executes compiled candidates and returns a comprehensive feedback object $\mathcal{F}_c$, encapsulating validation metrics, mathematical probes, user diagnostic reports, and consumed computational costs. The Distiller then analyzes the hypothesis $\mathcal{H}_c$, the code delta $\Delta_c$, and the empirical feedback $\mathcal{F}_c$ to synthesize mechanism-level lessons $\mathcal{D}_c$. These lessons explicitly record the tested mechanism, supporting evidence, applicable conditions, and failure attributes.

\paragraph{Search control and active co-adaptation.} 
After evaluating $v_c$, the Pruner acts as a search router rather than a simple filter. Jointly analyzing quantitative metrics and the Distiller's lessons $\mathcal{D}_c$, the Pruner dynamically dictates branch state transitions via structured outputs. First, for paths with irrecoverable collapses, it assigns a \texttt{[Pruned]} state, masking them from the Operator Selector to conserve computation. Second, to address cascaded pipelines, it enables active downstream co-adaptation. When an upstream innovation (e.g., Tokenizer edit) causes a metric regression due to cross-stage incompatibility, the Pruner emits a \texttt{[Needs-Downstream-Adaptation]} tag. Subsequent Operator Selector prompts are strictly conditioned on these tags, forcing the system to freeze upstream parameters and mutate the downstream architecture. Consequently, tightly coupled cross-stage mechanisms are co-evolved rather than prematurely discarded.

\section{Experiments} \label{sec:exp}
In this section, we conduct numerical experiments to answer five research questions.
\textbf{(RQ1)}: Does code-level \emph{DAG} evolution with domain-specific feedback beat both general and recommender-specific evolutionary frameworks?
\textbf{(RQ2)}: Does \emph{stage-aware cost planning} overcome \emph{cost blindness} and budget for expensive upstream (Tokenizer) innovations?
\textbf{(RQ3)}: Do \emph{active downstream co-adaptation} and \emph{cross-branch merging} overcome \emph{short-sighted evaluation}?
\textbf{(RQ4)}: Can the evolved architecture transfer to real-world industrial scenarios?
\textbf{(RQ5)}: Are user diagnosis and mathematical probes \emph{causally} useful in driving mutations?

\paragraph{Setup and Protocol.}
All frameworks evolve on a unified platform under a fixed budget $B=24$ GPU-hours, using Claude CLI~\citep{liu2026diveclaudecodedesign} with all inference routed to GLM-5.2~\citep{glm5team2026glm5vibecodingagentic}, starting from an identical TIGER seed codebase (plus post-training variants SDPO, SPRec, ReReGRPO, APAO) on the Amazon Office Products dataset. The final champion architecture evolved on this public dataset is subsequently transferred to Tencent's large-scale industrial platform for real-world validation (Section~\ref{sec:real-world}). We compare against general frameworks (AlphaEvolve~\citep{novikov2025alphaevolve}, DeepEvolve~\citep{liu2025deepevolve}, CORAL~\citep{qu2026coral}, Arbor~\citep{jin2026toward}) and recommender-specific ones (EvoRec~\citep{mu2026evorec}, Self-EvolveRec~\citep{kim2026self}); detailed configurations of the datasets, codebases, and baselines are provided in Appendix~\ref{appendix:dataset}, \ref{appendix:codebase}, \ref{appendix:baseline}. The platform exposes three feedback channels: \textbf{(i)} offline metrics (Hit@$K$/NDCG@$K$), \textbf{(ii)} deterministic mathematical probes, and \textbf{(iii)} LLM-based user diagnosis (Appendix~\ref{appendix:diagnostics}). Standard general baselines receive only offline metrics channel (i), whereas recommender-specific frameworks (EvoRec, Self-EvolveRec) and \method access all three channels. To strictly decouple the benefit of diagnostic feedback from evolutionary search topology, we further construct two naive integration baselines, namely \textbf{CORAL + Diag.} and \textbf{Arbor + Diag.}, which augment the strongest general search frameworks with identical multi-channel feedback. Model selection uses offline validation metrics alone. Both user diagnostic reports and Agent4Rec simulations are driven by DeepSeek-v4-flash\citep{deepseekai2026deepseekv4highlyefficientmilliontoken} with a temperature of 0 to ensure determinism and reproducibility. Crucially, evolutionary user diagnosis samples 50 simulated users, whereas post hoc Agent4Rec metrics~\citep{zhang2024agent4rec} are evaluated on 100 separate simulated users, keeping the two user cohorts strictly disjoint and non-overlapping to avoid evaluation leakage (Appendix~\ref{appendix:metrics}).

\paragraph{Main Results.} \label{sec:main-results}
As shown in Table~\ref{tab:main}, we make the following observations.
\textbf{(1) Consistent dominance.} \method leads on every metric, improving over the strongest non-evolutionary baseline APAO by $+24.7\%$/$+28.5\%$ on Hit@10/@20 and $+37.4\%$/$+37.2\%$ on NDCG@10/@20.
\textbf{(2) Disentangling domain feedback from search topology.} A critical question is whether the gains of \method stem simply from richer feedback or from its DAG-based evolutionary mechanisms. Comparing Arbor with Arbor + Diag. confirms that domain diagnostics indeed help (Hit@10 increases from $0.0792$ to $0.0814$). However, naively integrating domain feedback into general tree-based or population-based frameworks remains suboptimal: \method still significantly outperforms Arbor + Diag. ($+6.8\%$ Hit@10, $+16.9\%$ NDCG@10) and CORAL + Diag. ($+8.1\%$ Hit@10, $+16.0\%$ NDCG@10). This empirical gap demonstrates that feedback richness alone cannot substitute for structural search capabilities: without DAG-based cross-branch merging and stage-aware cost planning, general search frameworks fail to afford upstream tokenizer retraining or overcome temporary regressions during downstream co-adaptation. Furthermore, \method beats the strongest domain-specific baseline Self-EvolveRec by $+21.0\%$ Hit@10 / $+32.8\%$ NDCG@10 due to its multi-branch exploration.
\textbf{(3) Offline and simulated experience move together.} Beyond offline metrics, the Agent4Rec leaps (View $+31.3\%$, Satisfaction $+19.7\%$, Depth $+25.4\%$ over APAO) confirm that the discovered architectures yield genuinely better recommendations, answering \textbf{RQ1}.

\begin{table}[t]
\centering
\small
\setlength{\tabcolsep}{3.2pt}
\renewcommand{\arraystretch}{1.12}
\resizebox{0.9\columnwidth}{!}{%
\begin{tabular}{l cccc ccc}
\toprule
& \multicolumn{4}{c}{\textbf{Testset Metrics} $\uparrow$} & \multicolumn{3}{c}{\textbf{Agent4Rec} $\uparrow$} \\
\cmidrule(lr){2-5}\cmidrule(lr){6-8}
\textbf{Method} & Hit@10 & Hit@20 & NDCG@10 & NDCG@20 & View & Sat. & Depth \\
\midrule
\multicolumn{8}{l}{\textit{$\blacktriangledown$ Existing Non-Evolutionary Baselines}}\\
TIGER (Pre-train)~\citep{rajput2023tiger} & 0.0626 & 0.0944 & 0.0316 & 0.0396 & 0.270 & 5.45 & 2.45 \\
SDPO~\citep{chen2024sdpo}          & 0.0614 & 0.0946 & 0.0313 & 0.0396 & 0.288 & 5.75 & 2.55 \\
SPRec~\citep{gao2025sprec}         & 0.0626 & 0.0975 & 0.0310 & 0.0398 & 0.305 & 5.95 & 2.70 \\
ReReGRPO~\citep{tan2025reinforced} & 0.0663 & 0.0991 & 0.0331 & 0.0414 & 0.333 & 6.20 & 2.80 \\
APAO~\citep{Yu_2026}               & 0.0697 & 0.1081 & 0.0348 & 0.0444 & 0.355 & 6.45 & 2.95 \\
\midrule
\multicolumn{8}{l}{\textit{$\blacktriangledown$ General Evolutionary Frameworks (Offline Metrics Only)}}\\
AlphaEvolve~\citep{novikov2025alphaevolve} & 0.0685 & 0.1062 & 0.0338 & 0.0440 & 0.345 & 6.30 & 2.85 \\
DeepEvolve~\citep{liu2025deepevolve}  & 0.0705 & 0.1110 & 0.0345 & 0.0465 & 0.360 & 6.52 & 2.95 \\
CORAL~\citep{qu2026coral}            & 0.0785 & 0.1250 & 0.0395 & 0.0520 & 0.405 & 7.05 & 3.35 \\
Arbor~\citep{jin2026toward}          & 0.0792 & 0.1235 & 0.0388 & 0.0535 & 0.412 & 6.98 & 3.42 \\
\midrule
\multicolumn{8}{l}{\textit{$\blacktriangledown$ Naive Integration (General Search + Domain Diagnostics)}}\\
CORAL + Diag.                        & 0.0804 & 0.1276 & 0.0412 & 0.0542 & 0.420 & 7.18 & 3.48 \\
Arbor + Diag.                        & 0.0814 & 0.1272 & 0.0409 & 0.0550 & 0.425 & 7.12 & 3.51 \\
\midrule
\multicolumn{8}{l}{\textit{$\blacktriangledown$ Recommender-side Evolutionary Frameworks (with Diagnostics)}}\\
EvoRec~\citep{mu2026evorec}        & 0.0692 & 0.1075 & 0.0340 & 0.0452 & 0.352 & 6.38 & 2.88 \\
Self-EvolveRec~\citep{kim2026self} & 0.0718 & 0.1165 & 0.0360 & 0.0482 & 0.375 & 6.68 & 3.08 \\
\midrule
\textbf{\method} (Ours) & \textbf{0.0869} & \textbf{0.1389} & \textbf{0.0478} & \textbf{0.0609} & \textbf{0.466} & \textbf{7.72} & \textbf{3.70} \\
\midrule
$\Delta$ vs. Arbor + Diag.
              & +6.8\%  & +9.2\%  & +16.9\% & +10.7\% & +9.6\%  & +8.4\%  & +5.4\% \\
$\Delta$ vs. Self-EvolveRec
              & +21.0\% & +19.2\% & +32.8\% & +26.3\% & +24.3\% & +15.6\% & +20.1\% \\
$\Delta$ vs. APAO
              & \textbf{+24.7\%} & \textbf{+28.5\%} & \textbf{+37.4\%} & \textbf{+37.2\%} & \textbf{+31.3\%} & \textbf{+19.7\%} & \textbf{+25.4\%} \\
\bottomrule
\end{tabular}%
}
\caption{Main results. Evolutionary methods evolve the complete GR pipeline, from upstream Tokenizer to downstream post-training. $\Delta$ rows show the relative improvement of \method over Arbor + Diag., Self-EvolveRec, and APAO.}
\label{tab:main}
\vspace{-2mm}
\end{table}

\begin{figure}[t]
\centering
\includegraphics[width=0.9\columnwidth]{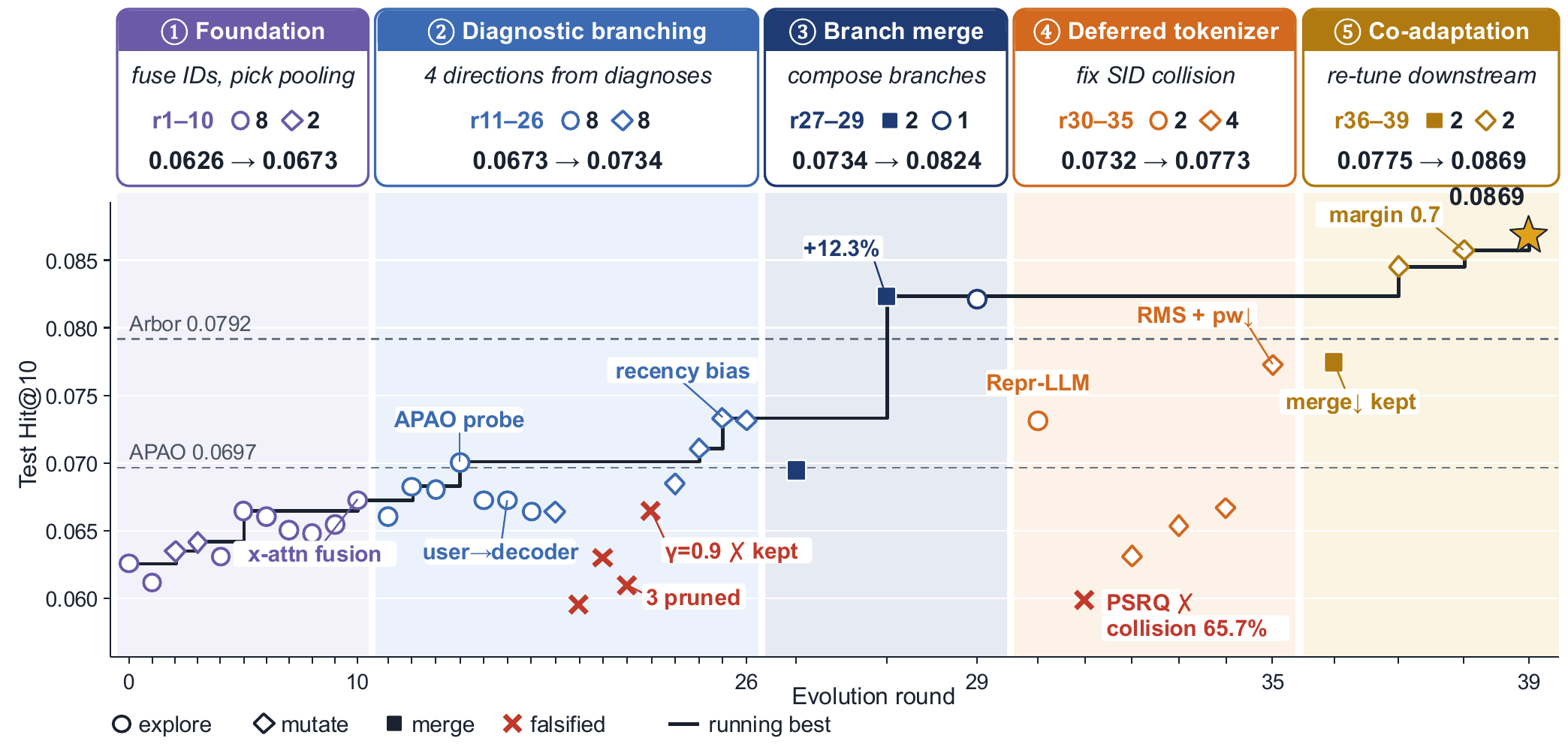}
\vspace{-2mm}
\caption{Evolution trajectory of \method (39 rounds, test Hit@10), where candidate models from every round are independently inferred offline on the test set post hoc to trace generalization without leakage. Header cards summarize the five phases (goal, rounds, operator counts, and Hit@10 change); marker shape denotes the mutation operator, red crosses represent falsified nodes, and the line traces the running best score. The x axis is scaled per phase.}
\label{fig:trajectory}
\vspace{-2mm}
\end{figure}

\paragraph{Analysis of Evolution Trajectory (RQ1-RQ3).} \label{sec:trajectory}
Figure~\ref{fig:trajectory} illustrates the 39-round evolution trajectory from the initial TIGER seed (Hit@10 $0.0626$) to the final architecture ($0.0869$) across five distinct phases, directly answering RQ1--RQ3. \textbf{(Phase 1)} The framework first establishes a robust baseline by injecting both collaborative and semantic IDs into the TIGER encoder, adjusting InfoNCE weights, and grid-searching feature fusion and user pooling methods ($0.0626\!\to\!0.0673$). \textbf{(Phase 2)} Next, \method opens multiple orthogonal branches based on diagnostic signals. This DAG structure safely explores and deepens parallel mutations (e.g., recency bias, user-conditional decoding) without destabilizing the main trunk, demonstrating a clear advantage over linear frameworks (RQ1). \textbf{(Phase 3)} It then pairwise merges these independently matured branches and stacks them with APAO, creating a single-step performance leap to $0.0824$ ($+12.3\%$) and validating the effectiveness of cross-branch merging (RQ3). \textbf{(Phase 4)} Because modifying the upstream tokenizer requires expensive full-pipeline retraining, \method strategically defers this to avoid budget waste. It tests the new Repr-LLM encoder on an isolated branch, where introducing PSRQ initially causes a collapse to $0.0599$. Diagnostics quickly repair it via RMS normalization and parameter sweeps before trunk integration, demonstrating how stage-aware cost planning overcomes cost blindness (RQ2). \textbf{(Phase 5)} Finally, when this repaired tokenizer is merged into the main trunk, performance initially regresses to $0.0775$. Instead of discarding it via greedy search, \method retains the node and actively co-adapts downstream components by re-tuning the fine-margin loss. This successfully recovers performance, pushing the model to its final peak of $0.0869$ after a final merge with the APAO branch, thereby overcoming short-sighted evaluation (RQ3).

\begin{figure}[h]
\centering
\includegraphics[width=0.9\columnwidth]{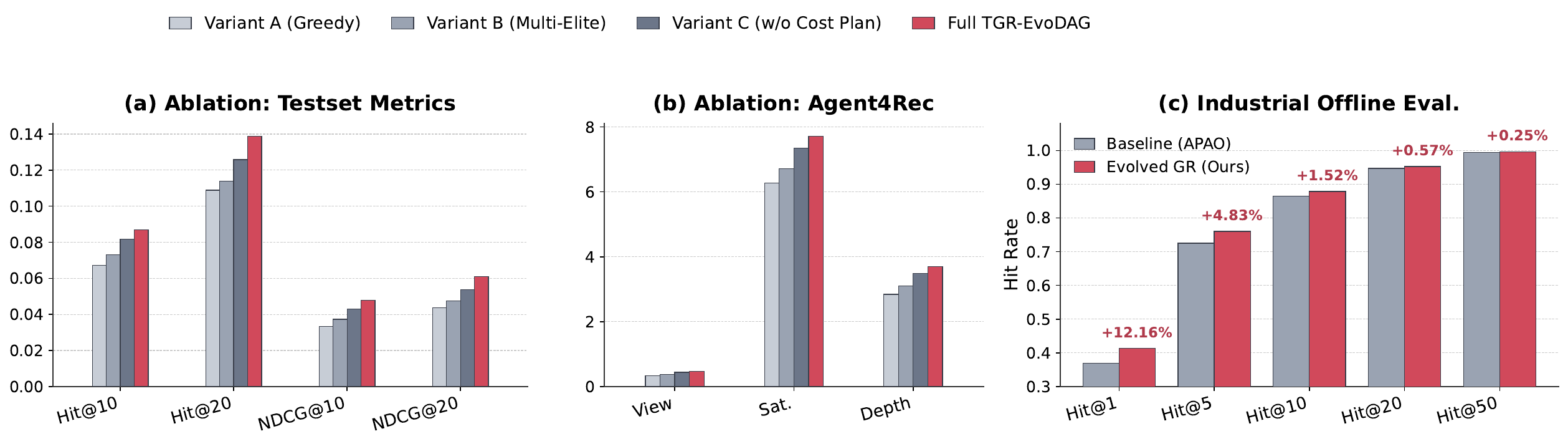}
\caption{Ablation of search mechanisms and industrial evaluation. (a) Testset
metrics and (b) Agent4Rec metrics under restricted search variants (A: greedy;
B: multi-elite; C: w/o cost planning) versus Full \method, where performance
rises monotonically from A to Full. (c) Real-world industrial offline evaluation
of the evolved GR pipeline over the APAO baseline across Hit@$K$.}
\label{fig:ablation}
\vspace{-2mm}
\end{figure}

\paragraph{Ablation of Search Mechanisms (RQ2 \& RQ3).} \label{sec:ablation}
We re-evolve under restricted variants: \textbf{A} (greedy: no multi-branch, prunes any regression), \textbf{B} (multi-elite but no cross-branch merge), and \textbf{C} (full search but without cost planning, uniform budget across stages). As shown in Figure~\ref{fig:ablation}(a,b), performance rises monotonically from A to Full \method on both testset and Agent4Rec metrics. \textbf{A} stalls at a local optimum, as its short-sighted greedy pruning cannot survive the temporary regressions required for downstream co-adaptation. \textbf{B} tolerates these delayed payoffs but remains blind to the compounding gains of merging independent branches, answering RQ3. \textbf{C} adds merging but suffers from cost blindness; lacking stage-aware planning, it exhausts the budget uniformly and cannot afford the expensive upstream tokenizer retraining. Full \method finally triages its budget properly to conquer the upstream stage, unlocking the terminal peak and answering RQ2.

\paragraph{Real-World Industrial Deployment (RQ4).} \label{sec:real-world}
We deploy the champion GR architecture to Tencent's commercial media platform with over 10 million daily active users. This deployment is carried out through Tencent's internal Uppimo platform, into which \method has been integrated as its core evolution engine. The production system follows a standard retrieval-pre-ranking-ranking cascade.
\textbf{Offline Transferability.} We train the champion architecture and the strongest baseline APAO from scratch using 10 days of industrial logs ($>$10 million interactions). As shown in Figure~\ref{fig:ablation}(c), our evolved model strictly improves over APAO across all metrics($+12.16\%$ Hit@1, $+4.83\%$ Hit@5, $+0.25\%$ Hit@50).
\textbf{Online A/B Testing.} We subsequently deployed the evolved model as a novel retrieval channel for a 14-day A/B test on 5\% of live traffic (serving $>$1M users). To ensure strict statistical rigor, we applied CUPED variance reduction and controlled the false discovery rate via tiered Benjamini-Hochberg correction ($q \le 0.05$). The online results reveal a significant overall system CTR increase of $+0.40\%$ ($p=0.019$). Furthermore, the new channel contributes over 10\% of unique exposures (relevant items completely missed by existing channels) and achieves a channel-specific CTR $\approx$13\% higher than the system average.

\paragraph{Validation of Diagnostic Signals (RQ5).} \label{sec:probe-validation}
To answer whether diagnostic signals are causally useful in driving mutations, we highlight three critical turning points where offline metrics alone offered no actionable direction. \textbf{(i) User Diagnosis:} When performance plateaued, simulated user feedback explicitly reported that recommendations were ``outdated,'' motivating the agent to inject an exponential recency bias. When an initial weak prior ($\gamma{=}0.9$) showed no offline improvement, diagnostics confirmed it was absorbed by learnable weights, prompting the agent to apply a stronger signal ($\gamma{=}0.5$) that successfully broke the plateau. \textbf{(ii) Mathematical Probes:} Standard metrics only indicate overall correctness, but mathematical probes isolated a specific ``same coarse, wrong fine'' error mode. This directly drove the agent to design a targeted within-coarse hard-negative contrastive loss that random mutation could not easily discover. \textbf{(iii) Distribution Probes:} When the new PSRQ tokenizer caused a catastrophic performance drop, an offline-only search would have discarded it. However, distribution probes pinpointed a semantic ID collision spike (to $65.7\%$) due to a k-means scale imbalance. This causal insight guided the agent to apply an RMS normalization repair, dropping collisions back to $1.0\%$ and unlocking the tokenizer's potential. Ultimately, while offline metrics only flag failures, these diagnostics provide the exact causal mechanisms required for the agent to execute targeted repairs.

\section{Conclusion} \label{sec:conclusion}

In this paper, we introduce \method, a multi-agent code evolution framework tailored for Generative Recommendation (GR). We observe that naively integrating existing evolution frameworks fails on GR’s cascaded pipelines due to two fundamental bottlenecks: cost blindness and short-sighted evaluation. To bridge this gap, \method harmonizes an advanced Code Directed Acyclic Graph (Code DAG) search mechanism with rich diagnostic feedback. Building upon this foundation, it introduces \textit{stage-aware cost planning} to explicitly triage edits based on actual GPU overhead, and \textit{active downstream co-adaptation} to safely navigate temporary regressions and unlock deferred gains. Extensive evaluations on a standardized GR platform demonstrate that \method achieves state-of-the-art performance, outperforming the leading non-evolutionary baseline (APAO) by up to 37.4\% in offline metrics and 19.7\% in simulated user satisfaction. In addition, successful migration to Tencent's real-world industrial scenarios yields a 12.16\% relative improvement in offline Hit@1 and a statistically significant 0.40\% gain in online Click-Through Rate. Moving forward, building on its integration into Tencent's Uppimo platform, we plan to directly evolve GR pipelines within complex industrial scenarios and to open the platform for public trial. Moreover, since Large Language Model (LLM) training inherently involves cascaded stages, we envision our framework can be naturally extended to optimize multi-stage LLM pipelines.

\clearpage

\subsection*{AI use statement}

In this work, we used generative AI tools only for tasks with recommended
disclosure. Specifically, (1) we used a large language model to assist in reading
and summarizing prior baseline works to aid our own understanding of the related
literature, and (2) we used a large language model to polish the writing (grammar,
phrasing, and clarity) of the manuscript. We did not use generative AI tools for
any tasks with required disclosure: the research ideas, the proposed \method
framework and its algorithms, the experimental design, and all code and
experiments were conceived and implemented by the authors. Data generation and
other required-disclosure tasks are not applicable to this work.

We have reviewed all AI-assisted work. All AI-generated summaries of baseline
methods were cross-checked by the authors against the original papers to ensure
factual accuracy, and all AI-polished text was manually verified by the authors
to preserve the intended technical meaning. We take full responsibility for the
final content of this work, including all text, claims, and artifacts.

\bibliography{iclr2027_conference}
\bibliographystyle{iclr2027_conference}

\clearpage

\appendix
\section{Appendix}

\subsection{Workflow}
\label{appendix:workflow}

Algorithm~\ref{alg:evodagrec} expands a single evolution round of \method. We present the
round in the same five stages as Section~\ref{sec:agents}, and use this section to make
precise what the main text states only qualitatively: the records exchanged between the
agents, the accounting of the budget $b^{(t)}$, and the transition of the branch status
$\mathcal{Q}$.

\paragraph{Agent records.}
Each agent is invoked as a single call constrained to one structured record, which the
harness parses directly; a response that cannot be parsed is discarded and the call is
repeated rather than repaired, so that no free-form residue enters the node attributes and
every intermediate decision remains machine-checkable. The Operator Selector consumes the
serialised graph $\mathcal{G}^{(t)}$ together with the remaining budget $b^{(t)}$ and returns
the operator $o_c$ and parent set $\mathcal{P}_c$; the Architect consumes these and the
ancestral lessons and returns the hypothesis $\mathcal{H}_c$; the Coder and Reviewer consume
$\mathcal{H}_c$ and return the codebase snapshot $\mathcal{B}_c$ with the delta $\Delta_c$;
the Distiller consumes $\mathcal{H}_c$, $\Delta_c$, and the feedback $\mathcal{F}_c$ and
returns the lesson $\mathcal{D}_c$; and the Pruner consumes $\mathcal{F}_c$, $\mathcal{D}_c$,
$o_c$, and the touched stage and returns the branch status $\mathcal{Q}_c$.
Two of these choices are deliberate. The Critic is instructed to reject a blueprint only
when it violates a hard invariant, cannot yield an executable forward pass, is a no-op, or
re-proposes a direction already falsified without addressing the recorded lesson
$\mathcal{D}_i$; a rejected blueprint costs a full round of agent calls, whereas an accepted
but unpromising one costs only a single evaluation, so the asymmetry favours acceptance. The
Distiller is given the Coder's workspace together with
$(\mathcal{H}_c, \Delta_c, \mathcal{F}_c)$ and inspects $\Delta_c$ before writing
$\mathcal{D}_c$, so that a lesson attributes the outcome to a mechanism (a specific layer,
objective, or initialization) instead of restating $\mathcal{H}_c$.

\paragraph{Budget accounting.}
A round is charged the GPU-hours measured by the environment, $\mathcal{C}_c$, and not the
expected cost estimated from the \texttt{target\_stage} and \texttt{training\_scope}
attributes. That estimate only informs the Operator Selector's strategic decision; the
decrement $b^{(t+1)} \leftarrow b^{(t)} - \mathcal{C}_c$ always uses the measured value,
which is recovered stage by stage and stored on the node, so that the cost profile available
to later rounds reflects the true stage-asymmetric overheads of the pipeline. A round that
fails review or is interrupted before yielding a usable candidate is charged
$\tau_{\text{fail}}$, the GPU-hours spent before the failure, so that repeatedly proposing
unexecutable candidates cannot be used to explore at no cost. Token consumption is logged
separately and does not draw on $b^{(t)}$.

\paragraph{Branch-status transition.}
A node that passes review and yields a valid checkpoint, whether or not it improves on its
parent, keeps its default status and remains selectable, since an immediate regression on an
upstream edit is precisely the signal that downstream co-adaptation is required. Only the two
tagged statuses alter the search: \texttt{[Pruned]} is assigned on an irrecoverable collapse
(unexecutable or protocol-violating code, a broken item$\leftrightarrow$code mapping, or a
metric collapse with no diagnostic route) and removes the subtree permanently, masking it from
the Operator Selector; \texttt{[Needs-Downstream-Adaptation]} is assigned when a metric
regression is caused by an upstream (tokenization or pre-training) edit and conditions the
subsequent Operator Selector prompts so that upstream parameters are frozen and the child node
must mutate the downstream stage to restore cross-stage compatibility.

\begin{algorithm}[t]
\caption{\method}
\label{alg:evodagrec}
\begin{algorithmic}[1]
\REQUIRE Seed codebase $\mathcal{B}_0$, maximum GPU-hour budget $B$, standardized environment $\mathcal{E}$, failure penalty $\tau_{\text{fail}}$
\ENSURE The best candidate node selected by validation performance
\STATE Initialize root node $v_0$ containing $\mathcal{B}_0$
\STATE Initialize global graph $\mathcal{G}^{(0)} = (\{v_0\}, \emptyset)$
\STATE $t \leftarrow 0$, $b^{(0)} \leftarrow B$
\WHILE{$b^{(t)} > 0$}
    \STATE \textcolor{gray}{\# 1. Stage-Aware Operator and Parent Selection}
    \STATE $(o_c, \mathcal{P}_c) \leftarrow \textsc{Selector}(\operatorname{Serialize}(\mathcal{G}^{(t)}), b^{(t)})$

    \STATE \textcolor{gray}{\# 2. Hypothesis Development}
    \STATE $\mathcal{H}_c \leftarrow \textsc{ArchitectCritic}(\mathcal{P}_c, o_c, \mathcal{G}^{(t)})$

    \STATE \textcolor{gray}{\# 3. Implementation and Review}
    \STATE $\mathcal{B}_c, \Delta_c \leftarrow \textsc{CoderReviewer}(\mathcal{H}_c, \mathcal{P}_c, o_c)$

    \STATE \textcolor{gray}{\# 4. Evaluation and Lesson Distillation}
    \IF{$\mathcal{B}_c$ passes review and is executable}
        \STATE $\mathcal{F}_c, \text{tokens}_t, \mathcal{C}_c \leftarrow \textsc{Evaluate}(\mathcal{B}_c, \mathcal{E})$ \hfill \textcolor{gray}{\# $\mathcal{C}_c$ is the actual GPU-hour cost}
    \ELSE
        \STATE $\mathcal{F}_c \leftarrow \text{Failure feedback}$, $\text{tokens}_t \leftarrow \text{Debugging cost}$, $\mathcal{C}_c \leftarrow \tau_{\text{fail}}$ \hfill \textcolor{gray}{\# measured GPU-hours spent before failure}
    \ENDIF
    \STATE $\textsc{Log}(\text{tokens}_t)$ \hfill \textcolor{gray}{\# Track LLM token consumption}
    \STATE $\mathcal{D}_c \leftarrow \textsc{Distill}(\mathcal{H}_c, \Delta_c, \mathcal{F}_c)$

    \STATE \textcolor{gray}{\# 5. Search Control and Active Co-Adaptation}
    \STATE Construct child node $v_c$ with attributes $(\mathcal{P}_c, o_c, \mathcal{H}_c, \Delta_c, \mathcal{B}_c, \mathcal{F}_c, \mathcal{D}_c, \mathcal{C}_c, \dots)$
    \STATE Add $v_c$ and derivation edges to form intermediate graph $\mathcal{G}'$
    \STATE $state, stage \leftarrow \textsc{Pruner}(\mathcal{F}_c, \mathcal{D}_c, o_c)$ \hfill \textcolor{gray}{\# $stage$: touched pipeline stage read from $\mathcal{H}_c$}
\IF{$state == \text{DEAD\_END}$}
    \STATE Mark $v_c$ as \texttt{[Pruned]}
\ELSIF{$state == \text{METRIC\_DROP} \text{ AND } \text{stage} == \text{UPSTREAM}$}
    \STATE Mark $v_c$ as \texttt{[Needs-Downstream-Adaptation]}
\ENDIF
\STATE $\mathcal{G}^{(t+1)} \leftarrow \text{UpdateGraph}(\mathcal{G}', v_c)$

    \STATE Update remaining budget: $b^{(t+1)} \leftarrow b^{(t)} - \mathcal{C}_c$
    \STATE $t \leftarrow t+1$
\ENDWHILE
\RETURN $\arg\max_{v_i \in \mathcal{V}} \operatorname{ValidationMetric}(\mathcal{F}_i)$
\end{algorithmic}
\end{algorithm}

\clearpage

\subsection{Datasets}
\label{appendix:dataset}
In this section, we introduce the open-source Amazon Office dataset used for evolution, as well as the industrial dataset used to verify whether the evolved GR architecture is effective in real-world industrial scenarios.

\textbf{Amazon Office Dataset}.
Following the evaluation protocol of the initial codebase~\citep{Yu_2026} and prior
generative recommendation studies~\citep{rajput2023tiger}, we conduct all evolution
experiments on the \textit{Office Products} category of the public Amazon Product
Review corpus\footnote{\url{https://jmcauley.ucsd.edu/data/amazon/links.html}}.
We \textbf{directly adopt the
preprocessed data released by}~\citet{Yu_2026}.
For completeness, we restate the protocol under which the released data were produced: each review is treated as one implicit positive interaction, and every user's interactions are sorted chronologically into a single behaviour sequence. 
Five-core filtering then iteratively discards users and items with fewer than 
five interactions. Each history is truncated to the most recent 20 items. 
Finally, the data is split with the standard leave-one-out strategy. For every user, 
the chronologically last interaction forms the test instance, the second-to-last 
forms the validation instance, and the rest are used for training.
All evolutionary frameworks therefore optimise and are selected on exactly the same
validation split, and the test split is touched only once, for the final checkpoint
selected within the budget $B$.
Item side information (title, brand, category, price and description) is concatenated
into a textual description and used exclusively by the item tokenizer.
The statistics of the released data are summarised in Table~\ref{tab:dataset_stats}.
Let $\mathcal{U}$, $\mathcal{I}$ and $\mathcal{D}$ denote the user set, the item set and
the interaction set, respectively. We report the average sequence length
$\overline{L}=|\mathcal{D}|/|\mathcal{U}|$ and the interaction density
$\rho = |\mathcal{D}| \big/ \left(|\mathcal{U}| \times |\mathcal{I}|\right)$.

\begin{table}[htbp]
  \centering
  \caption{Statistics of the Amazon Office dataset. We directly use the preprocessed
  data released by~\citet{Yu_2026}, obtained under five-core filtering and
  leave-one-out splitting.}
  \label{tab:dataset_stats}
  \begin{tabular}{lrrrrr}
    \toprule
    Dataset & $|\mathcal{U}|$ & $|\mathcal{I}|$ & $|\mathcal{D}|$
            & $\overline{L}$ & $\rho$ \\
    \midrule
    Amazon Office & 4{,}905 & 2{,}420 & 53{,}258 & 10.86 & 0.45\% \\
    \bottomrule
  \end{tabular}
\end{table}

We deliberately select Office as the \emph{evolution} dataset rather than a larger
benchmark. This is because a compact corpus keeps a single train/validate cycle
below one GPU-hour, so that the unified budget of $B=24$ GPU-hours admits a
meaningful number of evolution rounds for \emph{every} framework.

\textbf{Real-world Industrial Dataset}.
The industrial data are collected from Tencent's large-scale commercial media platform with over 10 million daily active users. The production model is trained continuously on streaming logs. For a controlled offline comparison, all models are trained on the same 10 consecutive days of logs (over 10 million interactions on more than 300 thousand items) and evaluated on the following day. A positive interaction is a click. The item text fed to the tokenizer concatenates textual metadata (e.g., title, category, and tags) with a caption of the item image generated by a multimodal LLM. User histories are truncated to the 500 most recent interactions. The platform serves recommendations through a retrieval--pre-ranking--ranking cascade, in which the evolved GR model is added as one retrieval channel for the online A/B test (Section~\ref{sec:real-world}). Exact user and item counts and absolute CTR values are withheld for confidentiality.

\subsection{GR Codebase}
\label{appendix:codebase}
In this section, we introduce the codebase provided for the evolutionary framework.

All frameworks evolve upon an identical seed codebase, which is built directly on the
officially released APAO implementation~\citep{Yu_2026} together with its preprocessed
data and its original tokenizer configuration.
The codebase is organised into four decoupled modules: (i) item tokenization, 
(ii) backbone definition, (iii) the pre-training objective, and (iv) a pluggable 
post-training objective. All four share one data loader and one beam-search evaluator.
What is held invariant across all frameworks is the \emph{environment}: the preprocessed
interaction data, the train/validation/test split, the evaluation protocol and the budget
$B$. Everything above the data layer, including the tokenization module, is exposed as
mutable code and may be modified by a candidate during evolution. 

\textbf{Item Tokenization}.
Every item is represented by a sequence of discrete semantic tokens rather than a
single atomic ID. The seed codebase ships with the tokenizer configuration
of~\citet{Yu_2026}: a semantic embedding is first extracted for each item by feeding its
concatenated textual metadata into \textsc{Llama-3.1-8B-Instruct} and taking the
final-layer representation; a \textbf{4-level RQ-$K$-means} quantizer then residually
quantizes each embedding into $T=4$ codes $(y^1,y^2,y^3,y^4)$, one per level. Each level
owns an independent codebook, and a de-duplication suffix is appended whenever two items
collide on all four levels, guaranteeing an injective item$\leftrightarrow$code mapping.
A user history $x=\{i_1,\dots,i_{n-1}\}$ is thus flattened into a token sequence of
length $4(n-1)$, and predicting the next item amounts to autoregressively generating
its $4$ semantic tokens.
Note that the tokenization module is \emph{not} frozen: a candidate is free to re-fit or
redesign the quantizer (e.g.\ changing the number of levels, the codebook size or the
quantization algorithm) as part of its proposed modification, in which case the
downstream backbone is retrained accordingly. Only the upstream interaction data and the
splits are fixed. The seed baselines reported in the main text all use the original
configuration above.

\textbf{Backbone}.
The seed backbone is TIGER~\citep{rajput2023tiger}, a T5-style Transformer
\emph{encoder-decoder} operating entirely in the semantic-token space. The encoder
consumes the flattened history tokens and the decoder autoregressively emits the $4$
codes of the target item; at inference time decoding is performed with constrained beam
search over the valid code trie. Both the encoder and the decoder consist of $4$ layers
with $6$ self-attention heads of head dimension $64$, ReLU activations in all
feed-forward blocks, and a learnable token embedding dimension of $128$, yielding
roughly $0.01$B parameters in total. The full architectural and optimisation
configuration is listed in Table~\ref{tab:codebase_config}; all values are inherited
from~\citet{Yu_2026} without modification.

\begin{table}[htbp]
  \centering
  \caption{Architecture and optimisation configuration of the seed GR codebase.
  These values are shared by the pre-training baseline and all post-training variants,
  and constitute the identical starting point from which every evolutionary framework
  departs.}
  \label{tab:codebase_config}
  \resizebox{\columnwidth}{!}{%
  \begin{tabular}{llc}
    \toprule
    Group & Configuration & Value \\
    \midrule
    \multirow{4}{*}{Tokenizer}
      & Semantic encoder            & \textsc{Llama-3.1-8B-Instruct} \\
      & Quantizer                   & RQ-Kmeans \\
      & \# Levels $T$               & 4 \\
      & Collision handling          & de-duplication suffix \\
    \midrule
    \multirow{6}{*}{Backbone (TIGER)}
      & Architecture                & Transformer encoder-decoder \\
      & \# Encoder / decoder layers & 4 / 4 \\
      & \# Attention heads          & 6 \\
      & Head dimension              & 64 \\
      & Token embedding dimension   & 128 \\
      & Activation                  & ReLU \\
    \midrule
    \multirow{8}{*}{Optimisation}
      & Optimizer                   & AdamW \\
      & Peak learning rate          & $5\mathrm{e}\text{-}4$ \\
      & LR schedule                 & cosine decay, 1\% warmup \\
      & Batch size                  & 1024 ($128\times8$ grad.\ accum.) \\
      & Max epochs                  & 200 \\
      & Early-stopping patience     & 20 \\
      & Negative sampling           & uniform random \\
      & \# Negatives per positive   & 100 \\
    \midrule
    \multirow{3}{*}{Inference}
      & Decoding                    & constrained beam search \\
      & Beam size $K$               & 20 \\
      & Max history length          & 20 items \\
    \bottomrule
  \end{tabular}}
\end{table}

\textbf{Pre-training Baseline}.
\textbf{TIGER (Pre-train)}~\citep{rajput2023tiger} optimises the standard token-level
cross-entropy (maximum-likelihood) objective under teacher forcing,
\begin{equation}
  \mathcal{L}_{\mathrm{CE}}
  = -\sum_{t=1}^{T}\log \pi_\theta\!\left(y^{t}_{+}\mid x,\, y^{<t}_{+}\right),
\end{equation}
where $y_{+}$ denotes the ground-truth code sequence. Every post-training variant below
is initialised from this converged checkpoint, so that the observed differences are
attributable purely to the post-training objective.

\textbf{Post-training Variants}.
The seed codebase ships with four post-training objectives, which together span the
mainstream design space of preference- and reinforcement-based alignment for generative
recommendation. They constitute the algorithmic prior that the evolutionary frameworks
may recombine, ablate or extend.

\begin{itemize}[leftmargin=*]
  \item \textbf{SDPO}~\citep{chen2024sdpo} generalises Direct Preference Optimization to
  a multi-negative, softmax-weighted ranking loss. Given a positive item $e_{i,+}$ and a
  sampled negative set $\mathcal{N}$, it optimises
  \begin{equation}
    \mathcal{L}_{\mathrm{SDPO}} = -\log\sigma\!\left(
      -\log\!\!\sum_{j\in\mathcal{N}}\!
      \exp\!\Big(\beta\log\tfrac{\pi_\theta(e_{j,-}\mid x)}{\pi_{\mathrm{ref}}(e_{j,-}\mid x)}
      -\beta\log\tfrac{\pi_\theta(e_{i,+}\mid x)}{\pi_{\mathrm{ref}}(e_{i,+}\mid x)}\Big)
    \right),
  \end{equation}
  where $\pi_{\mathrm{ref}}$ is the frozen pre-trained reference policy. Crucially, the
  preference signal is defined at the \emph{complete sequence} level, i.e.\ on the
  cumulative likelihood of the whole item, with a single summation over $\mathcal{N}$.

  \item \textbf{SPRec}~\citep{gao2025sprec} is a self-play framework designed to
  suppress the over-recommendation / filter-bubble bias that plain DPO induces. Each
  self-play iteration performs one SFT step on the offline interactions followed by one
  DPO step in which the \emph{model's own predictions from the previous iteration} serve
  as negatives. Because the negative distribution is the current policy's own output,
  the DPO loss is implicitly re-weighted by the model logits, adaptively penalising
  items the model already over-exposes. No extra data or manual intervention is needed.

  \item \textbf{ReReGRPO}~\citep{tan2025reinforced} casts post-training as
  reinforcement learning with verifiable rewards on top of GRPO. Two domain-specific
  adaptations make GRPO viable in the constrained item space: (i) candidates are sampled
  by \emph{constrained beam search} rather than i.i.d.\ sampling, which yields a group of
  valid, distinct and on-policy hard negatives in a single pass; and (ii) the sparse
  binary rule reward (exact match with the target) is augmented with a \emph{ranking
  reward} that penalises a generated negative proportionally to its probability rank,
  $r(e_k)=R_{\mathrm{rule}}(e_k)+\lambda R_{\mathrm{rank}}(e_k)$. Advantages are
  normalised within the group,
  $\hat{A}_k=(r_k-\mathrm{mean}(\mathbf{r}))/\mathrm{std}(\mathbf{r})$, and a KL penalty
  towards the reference policy stabilises the update.

  \item \textbf{APAO}~\citep{Yu_2026} attacks the training-inference inconsistency
  directly by moving supervision from the sequence level to every intermediate
  \emph{prefix}. Defining the cumulative score of a length-$m$ prefix as
  $s^{m}=\sum_{t\le m}\log\pi_\theta(y^{t}\mid x,y^{<t})$, the pairwise variant optimises
  \begin{equation}
    \mathcal{L}_{\mathrm{pair}}
    = \mathcal{L}_{\mathrm{CE}}
    + \beta\sum_{m=1}^{T} w_m\Big[-\log\sigma\big(-\log\!\!\sum_{j\in\mathcal{N}_m}\!\exp(s^{m}_{j,-}-s^{m}_{i,+})\big)\Big],
  \end{equation}
  while the pointwise variant replaces the bracketed term with the negative
  length-normalised prefix log-likelihood. The prefix weights
  $\mathbf{w}\in\Delta^{T-1}$ are \emph{not} fixed but follow an exponentiated-gradient
  update, $w_m\!\leftarrow\!w_m\exp(\eta\mathcal{L}_m)$ followed by re-normalisation,
  which realises \emph{adaptive worst-prefix optimisation}: the step whose prefix is
  currently most likely to fall out of the beam receives the largest weight. 
\end{itemize}

\subsection{Evolution Baselines}
\label{appendix:baseline}

We compare \method against six evolutionary frameworks, grouped into
\emph{general evolutionary frameworks} (domain-agnostic) and
\emph{recommender-specific frameworks} (built for recommendation). Every baseline is
instantiated on the same seed codebase, executed by the same coding agent
(Claude CLI), and constrained by the same budget $B=24$ GPU-hours described in the main
text, so that the differences reported stem from the search strategy itself rather than
from the environment.

\paragraph{General evolutionary frameworks.}
\begin{itemize}[leftmargin=*]
  \item \textbf{AlphaEvolve}~\citep{novikov2025alphaevolve} orchestrates an autonomous
  LLM-driven evolutionary pipeline that iteratively evolves and optimizes entire
  algorithmic codebases by directly editing and testing
  code against evaluator feedback. It generalizes the FunSearch/Eureka line of LLM-driven
  code evolution from single functions to whole codebases, but is guided primarily by
  scalar metrics and carries no recommendation-specific inductive bias.

  \item \textbf{DeepEvolve}~\citep{liu2025deepevolve} extends the evolutionary loop with
  retrieval-augmented generation (RAG): it retrieves external scientific knowledge (e.g.,
  academic papers from arXiv) to systematically inform hypothesis generation and codebase
  refinement, grounding the evolution in prior work rather than relying on the model's
  internal knowledge alone.

  \item \textbf{CORAL}~\citep{qu2026coral} is a multi-agent framework for open-ended
  discovery that replaces rigid control flow with long-running agents that explore,
  reflect, and collaborate through a shared persistent memory and asynchronous execution.

  \item \textbf{Arbor}~\citep{jin2026toward} is a general autonomous-research framework
  built on Hypothesis Tree Refinement : a persistent tree links hypotheses,
  artifacts, evidence, and distilled insights across time. A long-lived coordinator
  manages the global strategy over the tree, while short-lived executors implement and
  test individual hypotheses in isolated worktrees; returned results are propagated back
  into the tree as reusable lessons that refine the search frontier. 
\end{itemize}

\paragraph{Recommender-specific frameworks.}
\begin{itemize}[leftmargin=*]
  \item \textbf{EvoRec}~\citep{mu2026evorec} is a multi-agent framework that co-evolves
  both the recommendation model and the optimization methodology driving it. Four agents
  (an Orchestrator, a Research Agent, a Code Agent, and a Skill Evolver) run a dual-track
  loop: the Research and Code agents iterate the model each round, while the Skill
  Evolver periodically distills reusable methodology from a persistent memory of past
  experiments into a self-evolving skill library. It is domain-aware but is guided
  primarily by offline metrics, without the directional diagnosis.

  \item \textbf{Self-EvolveRec}~\citep{kim2026self} introduces \emph{directional
  feedback} for recommender evolution, pairing a \emph{User Simulator}  that emits
  natural-language critiques with a \emph{Model Diagnosis Tool}  that probes
  internal structure (e.g., embedding collapse, ranking margin) and interprets the
  numbers into a structured diagnosis. A \emph{Diagnosis Tool Co-Evolution} mechanism
  keeps the diagnosis tool compatible with the evolving model, and an evolutionary
  archive retains the codebase history. 
\end{itemize}

\subsection{Multi-dimension Diagnostics}
\label{appendix:diagnostics}

During evolution the environment produces, for every executed candidate, a structured
feedback bundle assembled from three complementary diagnostic channels, following the
directional-feedback design of Self-EvolveRec~\citep{kim2026self} (which pairs a
qualitative \emph{User Simulator} with a quantitative \emph{Model Diagnosis Tool}) and
specialising it to the 4-layer semantic-ID (SID) generative recommender.
The three channels are: \emph{(i)} offline validation metrics, giving a macro-level view
of ranking quality; \emph{(ii)} deterministic mathematical probes, exposing internal
failure modes that aggregate metrics cannot express; and \emph{(iii)} qualitative user
diagnosis, surfacing behaviour-level deficiencies in natural language.
While offline metrics in Channel~\emph{(i)} are available to every framework, diagnostic feedback in Channels~\emph{(ii)} and~\emph{(iii)} is provided exclusively to domain-specific methods, naive integration variants (Arbor + Diag. and CORAL + Diag.), and \method. For these methods, the diagnostic content is emitted \emph{stage-aware}: the bundle carries separate fields for the supervised-fine-tuning stage (\texttt{sft\_*}) and the post-training stage (\texttt{post\_*}), so that multi-stage diagnostics are produced only if a candidate actually modifies post-training; a single-stage candidate (e.g., pure TIGER) receives only the \texttt{sft\_*} fields.

\paragraph{Channel (i): Offline validation metrics.}
The environment computes standard ranking metrics (Hit@$K$ and NDCG@$K$, $K\in\{10,20\}$)
on the validation set via constrained beam search (beam size $20$), as formalised in
Appendix~\ref{appendix:metrics}. These scalars quantify \emph{how well} the candidate
ranks, but not \emph{why} it fails, and are therefore complemented by the two channels
below. In the feedback bundle the metrics appear both as the final-model value
(\texttt{offline\_val}) and as the per-stage values (\texttt{sft\_stage\_val},
\texttt{post\_stage\_val}) to enable stage-level credit assignment.

\paragraph{Channel (ii): Deterministic mathematical probes.}
The probe suite deterministically measures structural and decoding properties of the
trained model, and is grouped into four families summarised in
Table~\ref{tab:diagnostic_probes}. All values are computed by the environment on the
validation split; the baseline probe set is fixed and cannot be altered by any framework.
\paragraph{Probe extensibility.}
Although the baseline probe set is fixed, a candidate is permitted to implement and
upload \emph{additional} probe scripts to broaden its own diagnostic coverage. The
environment merges candidate-contributed metrics into the baseline under an
\texttt{agent.*} namespace and enforces an additive-only policy (candidate probes may not
overwrite baseline probes), so the official diagnostic surface remains identical across
frameworks while a framework may still monitor candidate-specific signals. The execution
status of any submitted probe is reported back in the \texttt{sft\_probe\_fn\_status} /
\texttt{post\_probe\_fn\_status} fields.

\begin{table}[htbp]
  \centering
  \caption{The four families of deterministic mathematical probes. Each probe yields one or more numeric values (indexed by SID layer or popularity tier) and targets a specific internal failure mode of generative recommendation.}
  \label{tab:diagnostic_probes}
  
  \small
  \renewcommand{\arraystretch}{1.2} 
  
  \begin{tabularx}{\columnwidth}{@{} l >{\raggedright\arraybackslash}X @{}}
    \toprule
    Representative probes & Diagnosed failure mode \\
    \midrule
    
    \multicolumn{2}{@{}l}{\textbf{Family: \texttt{sid\_generation}}} \\
    \quad \texttt{layer\_entropy}, \texttt{layer\_confidence} & per-layer decoding confidence; last-step or coarse-level collapse \\
    \quad \texttt{layer\_top5\_mass} & \\
    \quad \texttt{prefix\_diversity\_topk} & whether the top-$K$ list concentrates on few shared prefixes (i.e., SID utilisation) \\
    \quad \texttt{branching\_factor\_per\_layer} & fan-out per SID layer; whether the top-$K$ list is a rich subtree or $K$ parallel trajectories \\
    \quad \texttt{avg\_lcp\_depth\_topk} & beam collapse to a single coarse prefix \\
    \quad \texttt{beam\_prefix\_collapse\_rate} & \\
    \midrule
    
    \multicolumn{2}{@{}l}{\textbf{Family: \texttt{sid\_generation\_raw}}} \\
    \quad \texttt{raw\_layer\_entropy}, \texttt{raw\_layer\_confidence} & intrinsic (unmasked) decoding behaviour, decoupled from trie constraints \\
    \quad \texttt{raw\_layer\_top5\_mass} & \\
    \quad \texttt{raw\_gt\_layer\_rank}, \texttt{raw\_gt\_layer\_prob} & whether the ground-truth SID token is buried below distractors \\
    \midrule
    
    \multicolumn{2}{@{}l}{\textbf{Family: \texttt{output\_behavior}}} \\
    \quad \texttt{repetition\_rate\_topk} & failure to suppress already-owned items \\
    \quad \texttt{off\_genre\_rate} & over-broad generalisation / monotone or scattered lists \\
    \quad \texttt{subcategory\_diversity\_topk} & \\
    \quad \texttt{pop\_bias\_corr\_pool}, \texttt{topk\_pop\_split} & popularity bias \\
    \quad \texttt{avg\_topk\_pop\_rank\_pct} & \\
    \quad \texttt{brand\_concentration\_topk} & brand over-fixation \\
    \midrule
    
    \multicolumn{2}{@{}l}{\textbf{Family: \texttt{representation\_health}}} \\
    \quad \texttt{embedding\_collapse\_score} & representation degeneration of item/token embeddings \\
    \quad \texttt{token\_embedding\_collapse.\{a,b,c,d\}} & per-position representation degeneracy \\
    
    \bottomrule
  \end{tabularx}
\end{table}

The raw numeric values are not fed to the coding agent directly. Instead, an
LLM-based \emph{Diagnosis Analyst} interprets them into a structured diagnosis following
the schema of Table~\ref{tab:standard_diagnosis}; the prompt instructs the analyst to
ignore any conclusion-inducing wording, to ground every statement in a named metric and
its value, and to populate the \texttt{headline\_metrics} field with the 3-6 numeric
values that most directly support the verdict (making the diagnosis auditable).

\paragraph{Channel (iii): Qualitative user diagnosis.}
Following the user-simulator paradigm in Self-EvolveRec~\citep{kim2026self}, we instantiate an LLM-based simulator to produce behaviour-level critiques. We construct a population of simulated users, each governed by a fixed persona spanning three traits (\emph{activity}, \emph{conformity}, and \emph{diversity}, each ranging from LOW to HIGH). Conditioned on their individual interaction histories, these simulated users evaluate the candidate's top-$K$ recommendations via a two-step protocol:

\begin{itemize}[leftmargin=*, topsep=2pt, itemsep=2pt]
    \item \textbf{Item-level Assessment:} The simulator inspects each recommended item to predict the user's reaction. It outputs a binary verdict (align/misalign) along with a short causal reasoning for the decision.
    \item \textbf{List-level Feedback:} Building on the item verdicts, the simulator abstracts the overall quality of the top-$K$ list. It identifies which decision signals the model correctly captured, which patterns were overgeneralized, and which user intentions were missed.
\end{itemize}

Finally, to mitigate individual-user bias and surface systematic flaws, a global aggregator (prompted as a \emph{Lead Analyst}) synthesizes all per-user reports. This yields a single, macro-level system critique that strictly adheres to the unified \texttt{StandardDiagnosis} schema, serving as the qualitative counterpart to the mathematical probes.

\paragraph{Unified diagnosis schema.}
Both the mathematical and the qualitative channels converge on a single schema
(Table~\ref{tab:standard_diagnosis}), so that the coding agent consumes one uniform
diagnostic interface regardless of which channel produced the content.

\begin{table}[htbp]
  \centering
  \caption{The unified \texttt{StandardDiagnosis} schema shared by the mathematical and qualitative diagnostic channels.}
  \label{tab:standard_diagnosis}
  \renewcommand{\arraystretch}{1.2}

  \begin{tabularx}{\columnwidth}{@{} l l >{\raggedright\arraybackslash}X @{}}
    \toprule
    Field & Type & Semantics \\
    \midrule
    \texttt{status} & string & verdict in \{\texttt{CRITICAL}, \texttt{NEEDS\_IMPROVEMENT}, \texttt{STABLE}\} \\
    \texttt{alignment} & list[str] & behaviours where the model aligns with user choice logic \\
    \texttt{misalignment} & list[str] & behaviours where the model misreads user intent \\
    \texttt{overused\_signals} & list[str] & decision patterns the model relies on excessively \\
    \texttt{missing\_signals} & list[str] & signals the model under-weights or misses \\
    \texttt{primary\_failure\_mode} & string & the dominant failure mode at the top of the list \\
    \texttt{system\_interpretation} & string & one sentence on how the model decodes user intent \\
    \texttt{diagnostic\_signals\_to\_track} & list[str] & measurable signals to confirm or refute the diagnosis \\
    \texttt{headline\_metrics} & dict[str,float] & the numeric values backing the diagnosis (math channel) \\
    \bottomrule
  \end{tabularx}
\end{table}

As an illustration, the following  qualitative diagnosis was produced for a
pure cross-entropy TIGER candidate, demonstrating the concrete, behavioural language the agent receives:

\begin{lstlisting}
{
  "status": "NEEDS_IMPROVEMENT",
  "system_interpretation": "The system interprets user intent as broad category affinity and brand popularity, often ignoring ownership, specific needs, and behavioural nuance such as activity level.",
  "misalignment": [
    "It recommends items the user already owns or recently purchased.",
    "It overgeneralizes from a single broad category, recommending many items from the same subcategory." 
  ],
  "overused_signals": [
    "Heavy reliance on broad category similarity (e.g., 'Office & School Supplies').",
    "Overemphasis on brand popularity and brand-level co-occurrence." 
  ],
  "primary_failure_mode": "The top items often include redundant or non-essential items because the system overgeneralizes category and brand signals."
}
\end{lstlisting}

\subsection{Metrics}
\label{appendix:metrics}
In this section, we will introduce offline recommendation metrics (Hit and NDCG) as well as metrics defined in Agent4rec that simulate user experience.

\textbf{Offline Recommendation Metrics}.
Under the leave-one-out protocol each user $u$ has exactly one ground-truth next item
$i_u^{\ast}$. Given the ranked list $R_u$ of length $K$ produced by constrained beam
search, let $\mathrm{rank}_u\in\{1,\dots,K\}\cup\{\infty\}$ denote the position of
$i_u^{\ast}$ in $R_u$ ($\mathrm{rank}_u=\infty$ if it is not retrieved).

\emph{Hit@$K$} measures whether the ground-truth item appears in the top-$K$ list:
\begin{equation}
  \mathrm{Hit}@K = \frac{1}{|\mathcal{U}|}\sum_{u\in\mathcal{U}}
  \mathbb{I}\!\left(\mathrm{rank}_u \le K\right)
  = \frac{1}{|\mathcal{U}|}\sum_{u\in\mathcal{U}}
  \frac{\left|R_u^{K}\cap\{i_u^{\ast}\}\right|}{\left|\{i_u^{\ast}\}\right|}.
\end{equation}

\emph{NDCG@$K$} additionally rewards placing the ground-truth item near the top of the
list. With binary relevance $\mathrm{rel}_j=\mathbb{I}(R_u[j]=i_u^{\ast})$,
\begin{equation}
  \mathrm{DCG}@K(u)=\sum_{j=1}^{K}\frac{2^{\mathrm{rel}_j}-1}{\log_2(j+1)}
  =\frac{\mathbb{I}(\mathrm{rank}_u\le K)}{\log_2(\mathrm{rank}_u+1)},
  \qquad
  \mathrm{IDCG}@K(u)=\frac{1}{\log_2 2}=1,
\end{equation}
so that
\begin{equation}
  \mathrm{NDCG}@K=\frac{1}{|\mathcal{U}|}\sum_{u\in\mathcal{U}}
  \frac{\mathrm{DCG}@K(u)}{\mathrm{IDCG}@K(u)}
  =\frac{1}{|\mathcal{U}|}\sum_{u\in\mathcal{U}}
  \frac{\mathbb{I}(\mathrm{rank}_u\le K)}{\log_2(\mathrm{rank}_u+1)} .
\end{equation}
We report $K\in\{10,20\}$. Since the beam size equals $20$, the top-$20$ list is exactly
the beam-search output, and the top-$10$ list is its truncation.

\textbf{Agent4Rec Metrics}.
Offline recommendation metrics summarise a single next-item prediction and are therefore blind
to list-level behavioural effects such as over-concentration. To complement them, we
additionally report three behaviour-level metrics obtained from the LLM-based user
simulator Agent4Rec~\citep{zhang2024agent4rec}.
In this simulator each generative agent is initialised from a real user profile
(activity, conformity and diversity traits) and interacts with the evaluated recommender
\emph{page by page}: four items are displayed per page, the agent chooses which items to
view and rates them on a $1$-$5$ scale, and then decides whether to request the next
page or to exit, being forced to exit after at most five pages. A rating higher than $3$
is treated as a \emph{like}.
Let $\mathcal{A}$ denote the evaluation agent population, and for agent $a$ let $E_a$ be
the number of exposed items, $L_a$ the number of liked items, and $p_a\in[1,5]$ the page
index at which $a$ exits. We report
\begin{align}
  \bar{P}_{\mathrm{like}} &= \frac{1}{|\mathcal{A}|}\sum_{a\in\mathcal{A}}\frac{L_a}{E_a},
  &
  \bar{N}_{\mathrm{like}} &= \frac{1}{|\mathcal{A}|}\sum_{a\in\mathcal{A}} L_a,
  &
  \bar{N}_{\mathrm{exit}} &= \frac{1}{|\mathcal{A}|}\sum_{a\in\mathcal{A}} p_a .
\end{align}
Higher values indicate a better simulated user experience for all three metrics.
$\bar{P}_{\mathrm{like}}$ measures the precision of the exposed list at the behaviour
level, $\bar{N}_{\mathrm{like}}$ measures the absolute volume of satisfied interests a
session delivers, and $\bar{N}_{\mathrm{exit}}$ measures engagement depth: a larger exit page indicates the agent kept browsing rather than abandoning the session early.

\emph{Agent4Rec metrics are used strictly for evaluation.} They are computed
\textbf{only once, on the single checkpoint finally retained under the budget $B$}, and
serve exclusively to compare the frameworks in the main results table. They are never
computed for intermediate candidates, never exposed to any evolving framework, and never
used for model selection. Model selection relies solely on the offline validation metrics
defined above. Consequently, no framework can optimise these numbers directly, and they
function as a genuinely held-out behavioural measurement.

We stress that this evaluation simulator is \textbf{fully disjoint} from the user
simulator that supplies the third feedback channel during evolution. The latter follows
the qualitative-diagnosis design of Self-EvolveRec~\citep{kim2026self} and emits
natural-language diagnostic reports (Appendix~\ref{appendix:diagnostics}) rather than
scalar scores. The two components are isolated along two axes: (i) the user populations
are drawn from disjoint user sets, so no agent that participates in evolution-time
diagnosis is ever reused for final measurement; and (ii) their outputs are of different
kinds and consumed by different parties: diagnostic text is consumed by the coding 
agent during the search, whereas the three metrics above are consumed only by 
us, after the search terminates. This separation rules out any feedback path from the evaluation protocol back into the evolutionary process. Within the evaluation simulator itself, the agent population, the random seeds and the interaction templates are fixed once and shared by all frameworks, so the reported comparison is deterministic and paired.

\end{document}